# Circle-based Eye Center Localization (CECL)

Yustinus Eko Soelistio[1,2], Eric Postma[2], Alfons Maes[2]
Information System Department[1], Tilburg center for Cognition and Communication[2]
Universitas Multimedia Nusantara[1], Tilburg University[2]
Tangerang, Indonesia[1], Tilburg, The Netherlands[2]

**Abstract**

*The ability to automatically detect eye center locations in video images allows for estimating gaze direction. This, in turn, facilitates the study of human-computer interaction and behavioral analyses of social interactions. We propose an improved eye center localization method based on the Hough transform, called Circle-based Eye Center Localization (CECL) that is simple, robust, and achieves accuracy at a par with typically more complex state-of-the-art methods. The CECL method relies on color and shape cues that distinguish the iris from other facial structures. The circle enclosing the iris is localized by means of the Hough transform and the center of the iris is determined using the intensity level within the detected circle. The accuracy of the CECL method is demonstrated through a comparison with 15 state-of-the-art eye center localization methods against five error thresholds, as reported in the literature. The CECL method achieved an accuracy of 80.8% to 99.4% and ranked first for 2 of the 5 thresholds. It is concluded that the CECL method offers an attractive alternative to existing methods for automatic eye center localization.*

## 1. Introduction

The automatic localization of the centers of eyes is an active area of research [1-9, 13-23]. In theory, finding the centers of the only circular structures in facial images does not seem very hard. However, in practice eye center localization is a tough challenge due to the noisy nature of natural visual images (occlusion by eyelids, hair or glasses) and inappropriate illumination conditions.

There are two types of methods to localize the eye centers. The first type relies on specialized devices, such as infrared cameras [1], the second type makes use of standard video cameras, such as webcams [2-9, 13-23]. Although the use of specialized devices can give very accurate results, their application is often limited to unnatural situations. Hence, there is a need for eye center localization methods applicable to standard video cameras in visible light.

Two state-of-the-art examples of video-based eye-center localization methods are due to Timm and Barth (2011) [3] and Valenti and Gevers (2008) [5]. Their methods can achieve high accuracy, which means that the error, i.e., the distance between the estimated and true eye center, is small. Both methods use color characteristics of the pupil and the iris to determine the approximate eye center location. However both methods reported that they suffered from interference from visual contours such as occlusion by highlight and objects located outside the iris-sclera region (such as eyebrows, glasses, or hair). Similar types of interference are reported in the eye-gaze detection method proposed by Smith, Yin, Feiner, and Nayar (2013) [2].

The sensitivity to interference from adjacent visual contours may be overcome by adopting a model that relies on a shape cue for the irises. Circularity provides a unique cue to the irises, since no other visual structure in the face is circular. Several methods have been proposed using this idea, by combining edge detection [7, 8, 9] with the Hough transform (HT) [7-12]. Soltany, Zadeh, and Pourezza (2011; henceforth referred to as SZP) [8] proposed a two-step eye center localization method. In the first step, the horizontal and vertical intensity histograms of the image region containing the eye are used to determine the region of interest (approximate eye region). A dark image region (iris) reveals itself as minima in the two histograms. Using the Euclidean locations of the minima, the SZP method determines the location of the region of interest (ROI). In the second step, within the ROI the eye center is located by searching for a circle using the HT. The SZP method works well on clearly visible images of the eyes, but its first stage fails on images containing dark artifacts Figure 1 illustrates two failures of the SZP method.

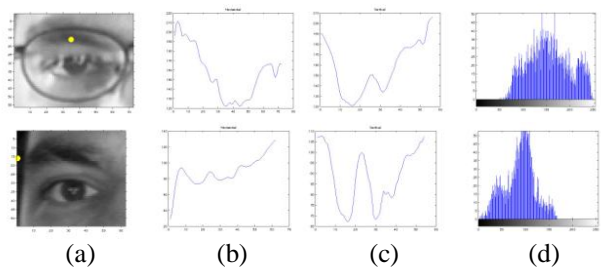

(a)      (b)      (c)      (d)

Figure 1. Illustration of failure of the SZP method. (a) Two images of the eye region containing dark artifacts (glasses and hair). (b, c) Horizontal and vertical histograms. (d) Image histograms. The small (yellow) dots in (a) shows the eye center estimates generated by the second stage of the SZP method.

This paper presents a new eye center localization method called the Circular-based Eye Center Localization (CECL) method that improves upon the SZP method by not suffering from dark artifacts.

The outline of the remainder of this paper is as follows. In Section II, we present the CECL method. Section III contains a comparative evaluation of the CECL method. Sections IV and V contain the general discussion and conclusions.

## 2. The CECL method

The CECL method consists of three stages, (1) eye region detection, (2) eye-region pre-processing and (3) eye center localization. In the first stage, the approximate eye locations are estimated by first localizing the face region and then localizing the eye region within the face region. In the second, pre-processing stage, potential disturbing visual elements (e.g., eyebrows, hair, or wrinkles) are removed.  Finally, in the third stage, circular objects are detected by means of the HT.  In some cases where HT fails to detect any circular object, CECL uses the center of the eye region as approximate eye center.

### 1.1. Eye localization

Eye localization proceeds in four steps. First, the image is smoothed with a Gaussian filter (size of $5 \times 5 px$, standard deviation = 5% of image width). Second the image region containing the face is detected. Third, the face region is cropped to 60% of its original height and the cropped image is cut into two equal halves along the vertical direction. Fourth, for each half, the eyes are detected. For both the face and eye detection steps, the Viola-Jones algorithm is used (Matlab's and mexopencv's implementations).

Whenever during the fourth step the eyes cannot be localized, the CECL method exhaustively searches for the most probable eye regions.

### 1.2. Pre-processing

The main purpose of the pre-processing stage is to remove unwanted artifacts that may hamper the HT-based circle detection. Pre-processing proceeds in four steps: (1) cropping, (2) contrast enhancement, (3) static binarization, and (4) morphological closing.

In the cropping step the top part of the image containing the eyebrow is removed. The size of the top part is defined by the parameter $T_e$.

In the second step the contrast of the cropped image is enhanced by means of histogram equalization.

The third step maps the equalized grayscale image onto a binary image by setting all pixels with a value $>T_b$ to 1 and to 0, otherwise. The optimal value of the threshold parameter $T_b$ is determined by means of machine learning (5-folds cross validation).  Ideally, the resulting binary image will show one or more full dark objects surrounded by a white background.  However, some white spots might appear in the middle of these dark objects.

In the fourth step, morphological closing is used to remove these white spots.

### 1.3. Eye center localization

In the third stage of the CECL method, consists of two steps: detecting the iris and finding its center. The circular HT [7-12] is used for detecting the circle enclosing the iris from the binary image. Given the resulting circular region, the minimum intensity value within the circular region is determined from the corresponding gray scale image.

Figure 2 provides an illustration of CECL's operation.

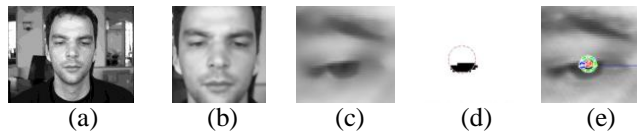

(a)  (b)  (c)  (d)  (e)

Figure 2. Illustration of the CECL method's operation. (a) Original image. (b) Result of face detection. (c) Result of eye detection. (d) Result of circle detection in binary image. (e) Eye center localization in the original image.

## 3. Performance evaluation

CECL's performance is evaluated on the BioID database [13], a challenging set of facial pictures taken under realistic conditions.  The database contains 1521 grayscale images ($384 \times 286\ pixels$) from 23 different subjects in different locations and under various illumination conditions.  Some of the subjects wear glasses and some subjects have curly hair obstructing their eyes.  They have different head pose and eye states.

We evaluate CECL in terms of accuracy by adopting the standard measure of normalized error ($e$) [13], which is defined as

$$e_{worst} = \frac{\max(d_l, d_r)}{\|C_l - C_r\|} \quad (1)$$

where $C_l$ and $C_r$ are the Euclidean positions of the true eye center locations of the left and right eyes, and $d_l$ and $d_r$ are the Euclidean distances between the predicted and real eye center locations, for the left and right eye respectively.

Timm and Barth [3] associated different error values as follows: $e = 0.25$ corresponds roughly to the distance between the eye center and the eye corner, $e = 0.10$ the diameter of the iris, and $e = 0.05$ the diameter of the pupil.  Predictions are considered good when $e \leq 0.05$.

We compare the CECL accuracy test result with the results from fifteen other methods. We also provide two additional normalized errors, the upper bound error, $e_{best} = \frac{\min(d_l, d_r)}{\|C_l - C_r\|}$, and the average error, $e_{average} = \frac{d_l + d_r}{2 \times \|C_l - C_r\|}$ [3, 5] to allow for comparison with other methods.

We implemented our method using Matlab R2012a and mexopencv on an Intel i5 2.5GHz multicore PC.  CECL was evaluated using 5-fold cross validation.

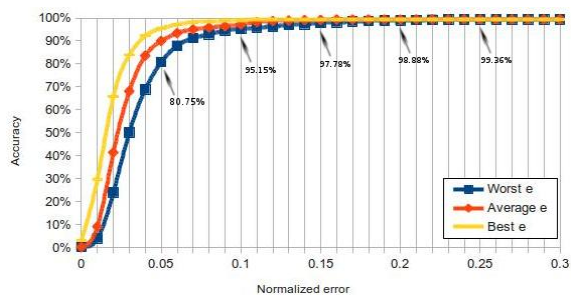

Figure 3. Normalized error of CECL method.

Figure 3 shows the accuracy of CECL for different $e$ thresholds. It shows that CECL can reach accuracy of 80.75% ($e \leq 0.05$), 95.15% ($e \leq 0.1$), and 97.78% ($e \leq 0.15$). The accuracy is relatively constant when $e > 0.15$.

Comparing CECL's performances with state-of-the-art methods, we adopt table 1 from Timm and Barth [3] extended with the evaluations reported in table 1 in Leo, Cazzato, De Marco, and Distante [23]. Adaptive methods are listed separately from non-adaptive methods to facilitate the comparison. Table 1 shows that CECL is outperforming all other methods when $e \leq 0.15$ and $e \leq 0.20$, and ranks second in all others except in $e \leq 0.05$ where it ranks third.

In terms of processing speed, the CECL method is somewhat slower than competitive methods. Figure 4 shows examples of successful (a-d) and unsuccessful (e-h) eye center localizations by the CECL method.

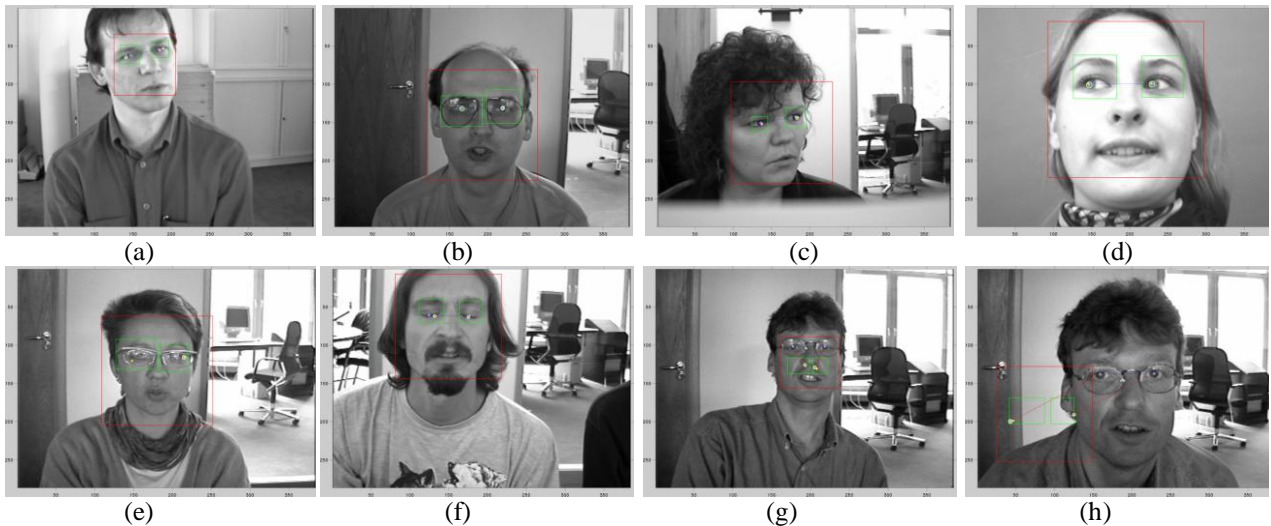

Figure 4. Samples of CECL eye center localization.

## 4. Discussion

The CECL method performs on a par with state-of-the-art eye center localization methods. The use of the HT to find circular objects appears to overcome the main limitation of image gradient methods [3], i.e., the interference of artifacts in the eye region (eyebrow, eyelid, and hair). CECL can increase the accuracy simply by limiting the ROI to the circular area determined using the HT.

Nevertheless, CECL's performance is restricted by its reliance on the presence of a visible circular feature of the iris. Specific illumination conditions can render the circular structure of the iris invisible, which gives rise to errors in CECL's adaptive binarization stage. The erroneous localizations in Figure 4(e-h) have various sources: strong reflection from glasses (e), thick eyelids during eye closure (f), incorrect eye localization (g), and incorrect face localization (h). These examples illustrate the two mean weaknesses of the CECL method: inaccurate face/eye detection and lack of "no detection" output. The inaccurate face and eye detections follow from the limitations of the employed Viola-Jones implementations. Future versions of CECL will be based on improved detectors. The lack of a "no-detection" output follows from CECL's implicit assumption that the iris is always visible. As a consequence, in cases where the iris is totally occluded, as in Figure 4(f), CECL will generate a false positive.

## 5. Conclusion

We proposed the CECL method, a simple and accurate eye center localization method using a HT-based circle detection. The CECL method performs at a level comparable with state-of-the-art methods, is less sensitive to interference from artifacts than gradient methods, is relatively simpler than other HT-based eye center localization methods, and operates successfully on low resolution images. By combining the shape and intensity cues, CECL can surpass other methods that rely on only either one cue.

Future research will address the two main weaknesses of the CECL method in an attempt to further boost eye-center localization performance.

We conclude by stating that the CECL method provides a viable approach to eye-center localization that deserves further study to achieve top level performance.

**Acknowledgments:** The authors would like to thank Bart Joosten for his valuable comments and suggestions.

Table 1. Comparison of CECL with other methods. (•) methods that do not use learning model. (S) method consider to be the current state-of-the-art. **Bold** value indicate best accuracy. (*) indicate second best accuracy.

| Method | $e \leq 0.05$ | $e \leq 0.10$ | $e \leq 0.15$ | $e \leq 0.20$ | $e \leq 0.25$ | Note |
|---|---|---|---|---|---|---|
| Asadifard and Shanbezadeh (2010) | 47.0% | 86.0% | 89.0% | 93.0% | 96.0% | • |
| Asteriadis et al. (2005) | 44.0% | 81.7% | 92.6% | 96.0% | 97.4% | • |
| Zhou and Geng (2004) | - | - | - | - | 94.8% | • |
| Timm and Barth (2011) | 82.5%* | 93.4% | 95.2% | 96.4% | 98.0% | •, S |
| Leo et al. (2014) | 80.7% | 87.3% | 88.8% | 90.9% | - | • |
| Kroon et al. (2008) | 65.0% | 87.0% | - | - | 98.8% | |
| Valenti and Gevers (2008) | **84.1%** | 90.9% | 93.8% | 97.0% | 98.5% | MIC+SIFT+kNN, S |
| Turkan et al. (2007) | 18.6% | 73.7% | 94.2% | 98.7%* | **99.6%** | |
| Campadelli et al. (2006) | 62.0% | 85.2% | 87.6% | 91.6% | 96.1% | |
| Niu et al. (2006) | 75.0% | 93.0% | 95.8% | 96.4% | 97.0% | |
| Chen et al., (2006) | - | 89.7% | - | - | 95.7% | |
| Hamouz et al. (2005) | 58.6% | 75.0% | 80.8% | 87.6% | 91.0% | |
| Christinacce et al. (2004) | 57.0% | **96.0%** | 96.5%* | 97.0% | 97.1% | |
| Behnke (2002) | 37.0% | 86.0% | 95.0% | 97.5% | 98.0% | |
| Jesorsky et al. (2001) | 38.0% | 78.8% | 84.7% | 87.2% | 91.8% | |
| **CECL** | 80.8% | 95.2%* | **97.8%** | **98.9%** | 99.4%* | |